\pdfoutput=1

\documentclass{article} 
\usepackage{Paper,times}

\usepackage{amsmath,amsfonts,bm}

\def\eqref#1{equation~\ref{#1}}

\def\1{\bm{1}}

\DeclareMathAlphabet{\mathsfit}{\encodingdefault}{\sfdefault}{m}{sl}
\SetMathAlphabet{\mathsfit}{bold}{\encodingdefault}{\sfdefault}{bx}{n}

\usepackage{hyperref}
\usepackage{url}
\usepackage{tcolorbox}
\usepackage{multirow}
\usepackage{cleveref}
\crefname{subsection}{§}{§§}
\usepackage{multicol}
\usepackage{tabularx,booktabs}
\usepackage{pdfpages}
\usepackage[ruled]{algorithm2e}
\usepackage{todonotes}
\usepackage{enumitem,kantlipsum}
\usepackage{graphicx}
\usepackage{subcaption}
\usepackage{adjustbox}
\usepackage{amsmath}

\usepackage[normalem]{ulem}

\title{SELF: Self-Evolution with Language Feedback}

\author{\textbf{Jianqiao Lu}$^1$\thanks{Leading co-authors with equal contribution.}\hspace{4px}\thanks{Work done during an internship at Huawei.}~, \textbf{Wanjun Zhong}$^{2*}$, \textbf{Wenyong Huang}$^{2*}$,\\
\textbf{Yufei Wang}$^2$\textbf{,}  \textbf{Qi Zhu}$^2$\textbf{,}  \textbf{Fei Mi}$^2$\textbf{,} \textbf{Baojun Wang}$^2$\textbf{,} \textbf{Weichao Wang}$^2$\textbf{,} \textbf{Xingshan Zeng}$^2$\textbf{,} \\
\textbf{Lifeng Shang}$^2$\textbf{,}  \textbf{Xin Jiang}$^2$ 
\& \textbf{Qun Liu}$^2$
\\
$^1$The University of Hong Kong \ \ \ $^2$Huawei Noah’s Ark Lab\\
\texttt{jqlu@cs.hku.hk, \{zhongwanjun1,wenyong.huang\}@huawei.com}\\
}

\newcommand{\method}{\textit{SELF}\xspace}

\iclrfinalcopy

\begin{document}
\maketitle
\begin{abstract}

Large Language Models (LLMs) have shown impressive adaptability in various fields, yet the optimal pathway of autonomous model evolution remains under-explored.
Drawing inspiration from the self-driven learning process of humans, we introduce \method (Self-Evolution with Language Feedback), a novel learning framework that empowers LLMs to continually self-improve their abilities. 
SELF initiates with a meta-skill learning process that equips the LLMs with capabilities for self-feedback and self-refinement. 
SELF employs language-based feedback for detailed and nuanced evaluations, pinpointing response flaws and suggesting refinements. 
Subsequently, the model engages in an iterative process of self-evolution: they autonomously generate responses to unlabeled instructions, refine these responses interactively, and use the refined and filtered data for iterative self-training, thereby progressively boosting their capabilities.
Moreover, the SELF framework equips the model with the ability to self-refine during inference, leading to further improved response quality.
Our experiments on mathematical and general tasks demonstrate that SELF enables the model to continually self-improve without human intervention. The SELF framework indicates a promising direction for the autonomous evolution of LLMs, transitioning them from passive information receivers to active participants in their development.

\end{abstract}

\section{Introduction}
Large Language Models (LLMs), like ChatGPT~\citep{chatgpt} and GPT-4~\citep{openai2023gpt4} , stand at the forefront of the AI revolution, demonstrating versatility across tasks. 
Despite their evident capabilities, the way towards achieving autonomous development of LLMs is still under-explored.

The development of automatically improved LLMs can draw inspiration from human self-driven learning mechanisms. 
When facing new challenges, humans naturally engage in a learning cycle of initial attempts, introspective feedback, and behavior refinement. This leads to a critical question: ``Can LLMs mimic the human learning process, utilizing self-refinement to enhance their inherent capabilities?''
Fascinatingly, {a recent study}~\citep{selfee2023} in top-tier LLMs such as GPT-4 has revealed emergent meta-skills for self-refinement, signaling a promising future direction for the self-evolution of LLMs. 
Despite this, current methods for LLM development typically rely on a  single round of instruction fine-tuning ~\citep{wei2021finetuned,zhou2023lima} with meticulously human-crafted datasets and reinforcement learning-based methods ~\citep{ouyang2022nips} that depend on an external reward model. 
These strategies not only require extensive resources and ongoing human intervention but also treat LLMs as mere passive repositories of information rather than active learners. 
These limitations hinder LLMs from tapping into their inherent capabilities, obstructing their progress toward a self-driven, autonomous learning paradigm.
\begin{figure}[t] 
\centering 
\includegraphics[width=\textwidth]{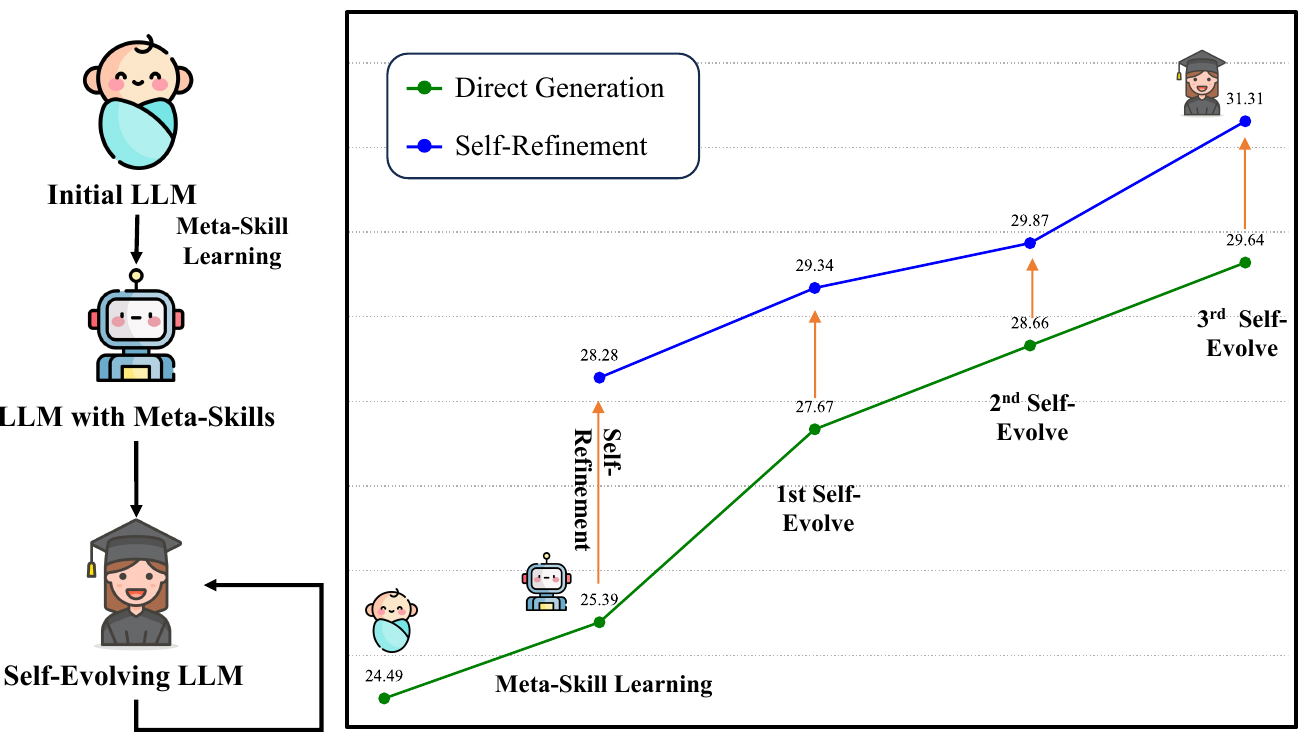}
\caption{Evolutionary Journey of SELF: An initial LLM undergoes successive self-evolution iterations (1st, 2nd, 3rd), enhancing its capabilities and acquiring a self-refinement meta-skill.}
\label{Fig:main_picture} 
\vspace{-0.2in}
\end{figure}

Thus, we introduce \method (Self-Evolution with Language Feedback) framework, designed to unlock the potential for autonomous self-evolution in LLMs. Figure \ref{Fig:main_picture} depicts how SELF mimics human-like self-driven learning, emphasizing progressive improvement of model capability with self-evolution training. 
At the core of SELF are the two meta-skills (\textit{self-feedback and self-refinement}), empowering the model to progressively self-evolve by training on its own synthesized data.
Additionally, SELF leverages self-generated natural language feedback to offer in-depth analysis and guidance for refining responses, without the need for external rewards or direct human guidance. 

Specifically, the SELF framework initiates by teaching LLMs essential meta-skills, namely self-feedback and self-refinement, using a limited set of examples. 
Once these skills are acquired, the model engages in a cycle of continuous self-evolution, iteratively training with extensive, self-generated data. 
Given a large-scale unlabeled corpus, this data is compiled from initial responses and refined through self-refinement and filtering, with model itself.
During this iterative process, the quality of self-evolution training data and model capability are interactively improved, fostering ongoing self-evolution of LLMs. 
Crucially, in the inference phase, these learned meta-skills enable LLMs to further enhance response quality via self-refinement.
In summary, the SELF framework transforms LLMs from passive recipients of data into active learners in self-evolution, and alleviates data scarcity issues by generating large-scale self-curated training datasets. This not only reduces the need for labor-intensive manual intervention but also promotes the continuous self-improvement of LLMs, establishing a more autonomous and efficient training approach.

We evaluate SELF in mathematical and general domains.
SELF notably improves the test accuracy on mathematical domains (6.82\% on GSM8k~\citep{cobbe2021gsm8k} and 4.9\% on SVAMP~\citep{pateletal2021nlp}), and increases the win rate on general domain (10\% on Vicuna testset~\citep{Vicunatestset} and 6.9\% on Evol-Instruct testset~\citep{xu2023wizardlm}), compared with typical supervised fine-tuning. 
There are several insights gained from our experiments.
Firstly, SELF can progressively enhance the model capability through self-evolution training.
Secondly, the learning of meta-skills, specifically self-feedback and self-refinement, is crucial not only for equipping the model with self-improvement abilities but also for boosting its direct response generation performance.
Finally, the model demonstrates further improvement in its responses through self-refinement during the inference stage.

The main contributions are summarized as follows:
(1) SELF empowers LLMs with self-evolving capabilities, allowing for autonomous model evolution, and reducing human intervention.
(2) SELF facilitates self-refinement into smaller LLMs, even with challenging math problems. The capability of self-refinement was previously considered an emergent characteristic of top-tier larger LLMs.
(3) Experiments demonstrate the effectiveness of SELF in both mathematical and general domains, confirming its advanced capabilities in self-evolution and self-refinement.

\section{Related Works}

\paragraph{Self-improvement in Inference}
Self-consistency~\citep{wang2022selfconsistency} is a straightforward and effective method to improve LLMs for reasoning tasks. 
After sampling a variety of reasoning paths, the most consistent answer is selected. 
During decoding, self-consistency is closely tied to the self-refinement capability of LLMs, on which our method is based. 
Unlike self-consistency, self-refinement applies to a broader range of tasks, going beyond reasoning tasks with unique correct answers.
{Various research efforts have been undertaken to enhance the output quality of LLMs through \emph{online self-improvement}~\citep{shinn2023reflexion,madaan2023self,selfee2023,chen2023teaching,ling2023deductive}.
The main idea is to generate an initial output with an LLM. Then, the same LLM provides feedback on its output and employs this feedback to refine its initial output. 
This process can be iterative until the response quality is satisfied.}
{While simple and effective, \emph{online self-improvement} necessitates multi-turn inference for refinement, leading to increased inference computational overhead. Most importantly, \emph{online self-improvement} does not prevent the model from repeating previously encountered errors, as the model's parameters remain unchanged.
In contrast, SELF can self-improve with evolution training. 

\paragraph{Autonomous Improvements of LLMs}
``Alignment'' \citep{leike2018scalable} aims to train agents to act in line with human intentions. 
Several research efforts~\citep{ouyang2022nips,bai2022training,scheurer2023training} leverage Reinforcement Learning from Human Feedback (RLHF) \citep{christiano2017nips}.
{RLHF begins with fitting a reward model to approximate human preferences.}
{Subsequently, an LLM is finetuned through reinforcement learning to maximize the estimated human preference of the reward model.}
{Reward Ranked Fine-tuning (RAFT) utilizes a reward model to rank responses sampled from an LLM. Subsequently, it fine-tunes the LLM using highly-ranked responses~\citep{dong2023raft}. }
Recent advancements in LLMs have explored Reinforcement Learning (RL) approaches that do not rely on human feedback. 
RLAIF~\citep{pang2023language} proposes to employ LLMs to label the preference data in replace of human supervision.
LLMs are updated progressively through online RL in interacting with the environment in~\citet{carta2023grounding}. 
{The connection between conventional RL research and RLHF in LLMs is discussed by~\citet{sun2023reinforcement}.
}
However, scalar rewards in Reinforcement Learning (RL) offer limited insights for evaluating complex reasoning tasks~\citep{lightman2023let}, as they fail to specify detailed errors and optimization paths. Recognizing this limitation, the SELF framework suggests utilizing natural language feedback, which effectively guides the self-evolution of LLMs. 
Unlike scalar rewards, which require a retrained model for each evaluation protocol and dimension, natural language feedback is more flexible, addressing multiple aspects simultaneously. 
Furthermore, the RLHF process is intricate and computationally intensive, relies on external reward models, and demands manual tuning of hyperparameters for optimal performance. This approach lacks the adaptability to evolve intrinsically with the model itself.

\begin{figure*}[ht] 
\centering
\includegraphics[width=\textwidth]{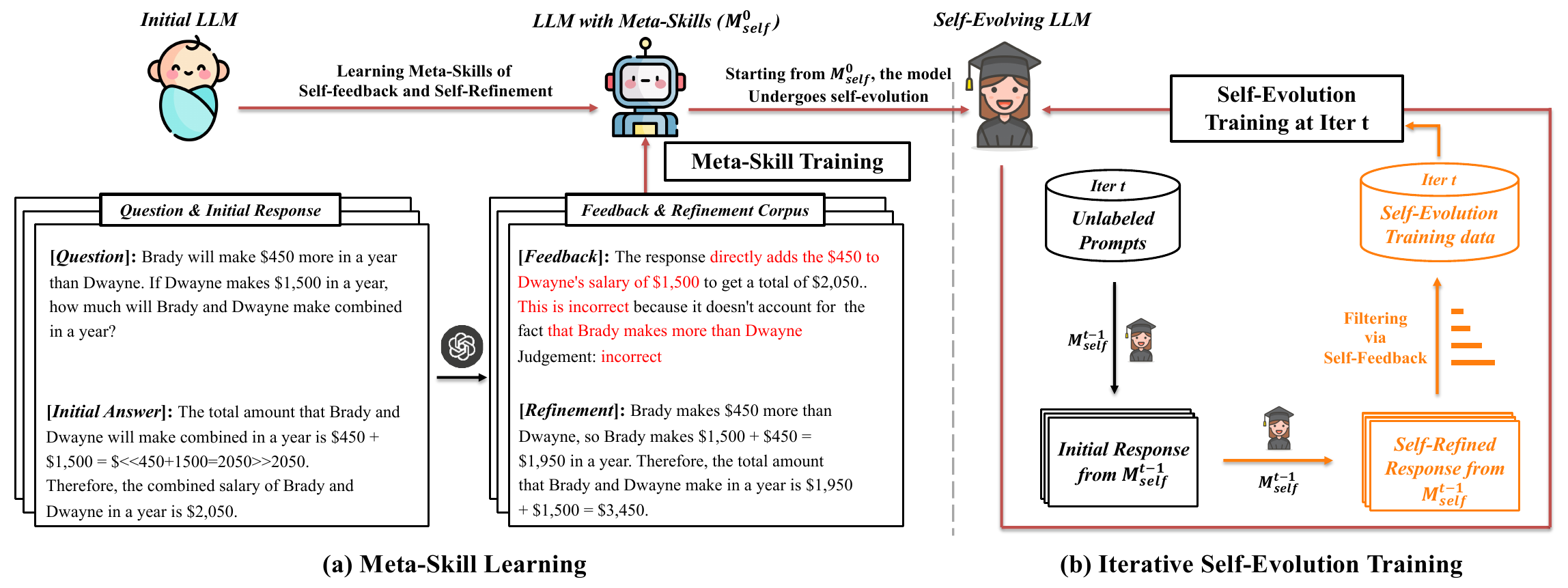}
\caption{Illustration of SELF. The ``Meta-Skill Learning" (left) phase empowers the LLM to acquire meta-skills in self-feedback and self-refinement. 
The (b)``Self-Evolution'' phase (right) utilizes meta-skills for self-evolution training with self-curated data, enabling continuous model enhancement.} 
\label{Fig:self-evolve-overview} 
\vspace{-0.15in}
\end{figure*}
\section{Method}
\label{sec:method}
{As depicted in Fig. \ref{Fig:self-evolve-overview}, the SELF framework enhances model capabilities through a two-stage learning phase:
(1) \textbf{Meta-skill Learning Phase}: This phase uses an annotated meta-skill training corpus to fine-tune the model, and equips the model with essential meta-skills for self-feedback and self-refinement with limited supervised examples, laying a foundation for self-evolution.
(2) \textbf{Self-Evolution Phase}:
With the acquired meta-skills, the model progressively improves through multiple
iterations of the self-evolution training process. 
Each iteration begins with the model itself autonomously creating high-quality training data from unlabeled prompts.
Then, the model is fine-tuned using this data.}
The process is further illustrated in Alg.~\ref{alg:self-evolution-process} in Appendix \ref{sec:algorithm}.

In SELF, \textbf{Natural Language Feedback} plays a crucial role in guiding the evolutionary process. This approach offers a more fine-grained and informative evaluation compared to the traditional method of using a  scalar reward. 
The latter evaluates quality along a single dimension with a numerical value from a reward model. In contrast, natural language feedback provides a detailed and descriptive analysis of the processes involved in a response, which is particularly critical in complex reasoning tasks. This also allows for evaluation across multiple dimensions, offering greater flexibility. Importantly, it guides the refinement process by suggesting directions for improvement.
The efficacy of natural language feedback in enhancing evaluation accuracy and model performance is shown in \cref{sec:exp-compare-rlhf}.


\subsection{Meta-Skill Learning} 
\label{sec:meta-skill}
Meta-skill learning targets on instill two essential meta-skills into LLMs for self-evolution.
(1) \textbf{Self-Feedback Ability}: This skill enables LLMs to evaluate their responses using natural language feedback. 
This provides the suggestion for further refinement, thus laying a solid foundation for subsequent self-refinement.
Self-feedback also enables the model to filter out low-quality self-evolution training data if a response is judged as unqualified by the model (\cref{sec:self-evolve-training-corpus}). 
(2) \textbf{Self-Refinement Ability}: 
 Self-refinement enables the model to optimize its responses based on self-feedback.
This ability has two applications:  (1) enhancing the quality of the self-evolution training corpus (\cref{sec:self-evolve-training-corpus}) and (2) improving model performance by refining the models' outputs during inference (\cref{sec:online-refinement}). 

These meta-skills are acquired by fine-tuning the model using the \textbf{Meta-Skill Training Corpus} (\cref{sec:meta-skill-corpus}) with designed training objective (\cref{sec:training-objective}). 
The resulting model is denoted as \(M_{\text{meta}}\).

\subsubsection{Meta-Skill Training Corpus}
\label{sec:meta-skill-corpus}
The meta-skill training corpus \(D_\text{meta}\) represents the generation, feedback, and refinement process.
It is constructed as follows:
(1) For each unlabeled prompt \(p\), the initial model \(M_{\text{init}}\) generates an initial response \(r\). 
(2) An annotator \(L\) provides evaluation feedback \(f\) for the initial response $r$, then produces a refined answer \(\hat{r}\) according to the feedback  \(f\).
Each instance in \(D_\text{meta}\) takes the form \((p, r, f, \hat{r})\), representing the process of response evaluation and refinement. 
An example instance of \(D_\text{meta}\) is provided in  \cref{app:meta-skill training corpus}.

\subsubsection{Training Objective}
\label{sec:training-objective}

In the meta-skill learning phase, the objective is to empower the initial model \( M_{\text{init}} \) to develop meta-skills, resulting in an enhanced model \( M_\text{meta} \).
This process is guided  by the cross-entropy loss following the maximum likelihood estimation (MLE) paradigm:  
\begin{align}
\label{eq:meta-skill learning}
\begin{aligned}
 \mathcal{L}_{\text{meta}}(\phi) = -   \mathbb{E}_{(p, r, f, \hat{r})  \sim D_\text{meta}}   
 \big[
 \log\tau_\phi(f|p, r)  +
  \log\tau_\phi(\hat{r}|p, r, f)  + 
  \beta\log\tau_\phi(\hat{r}|p)  \big],
\end{aligned}
\end{align}

where \(p\) is prompt,  \(r\) is the initial model response,  \(f\) is the feedback to the model response \(r\), and \(\hat{r}\) is the revised response based on feedback \(f\).
 \( \tau_\phi(y|x) \) denotes the probability distribution given by the auto-regressive language model with parameters \( \phi \) predicting the response \( y \) given the input prompt \( x \). 
 The coefficient \(\beta\) in~\cref{eq:meta-skill learning} controls a balanced emphasis on direct response generation and the model's capability for self-feedback and self-refinement.
 \paragraph{Insight.}

 Training with \(D_{\text{meta}}\) aims to achieve two goals: 
 (\romannumeral1) To guide the model in generating feedback (\(f\)) concerning its initial responses (\(r\)) (\textbf{self-feedback}) and subsequently employing this feedback to enhance the quality of the final answer (\(\hat{r}\)) (\textbf{self-refinement}).
 (\romannumeral2) 
 Training with \(D_{\text{meta}}\) instructs the model to process problems in a 
 Chain-of-Thought (CoT) manner. 
 This involves evaluating the initial response, integrating the feedback, and then revising the response in a chain process \(  \Psi(\hat{r} | p) := \sum_{r,f} \tau_\phi(r|p) \cdot \tau_\phi(f | p, r) \cdot \tau_\phi(\hat{r}|p, r, f)\).



\subsection{Self-Evolution Training Process}
The model \(M_{\text{meta}}\), equipped with meta-skills, undergoes progressive improvement through multiple iterations of the self-evolution training process. 
Each iteration of the self-evolution process begins with the model autonomously creating high-quality training data (\cref{sec:self-evolve-training-corpus}) from an unlabeled corpus.
With an unlabeled dataset of prompts, the model generates initial responses and then refines them through self-feedback and self-refinement. 
These refined responses, superior in quality, are further filtered with self-feedback and utilized as the training data for the model’s subsequent self-evolution training (\cref{sec:self-evolve-training}). 
This autonomous self-evolving process interactively improves LLMs as the improved model capability leads to better data quality, which in turn boosts model performance.
It also alleviates the data scarcity problem by self-generating data.

\subsubsection{Self-Evolution Training Data}
\label{sec:self-evolve-training-corpus}
 
Let \(M^{t}_{\text{evol}}\) denotes the model at  \(t^{th}\)  iteration and initialize \(M^{0}_{\text{evol}}\) from \(M_{\text{meta}}\).
During \(t^{th}\) self-evolution iteration ,
\(M^{t-1}_{\text{evol}}\) processes each unlabeled prompt \(p\) by first generating an initial response \(r\).
\(r\) is then refined using the model's self-feedback \(f\), resulting in a self-refined response \( \hat{r}\). 
The prompts and their corresponding self-refined responses\((p, \hat{r})\) are then incorporated into the \(t^{th}\) round self-evolution datasets, denoted as \(D^{t}_{\text{evol}}\), for subsequent self-evolution processes.


\textbf{Data Filtering with Self-feedback:} 
To enhance the quality of \(D^{t}_{\text{evol}}\), we employ the self-feedback capability of \(M^{t-1}_{\text{evol}}\) to filter out data of lower quality. 
\(M^{t-1}_{\text{evol}}\) evaluates the self-refined data, \( \hat{r}_{\text{evol}}\), keeping only the responses that meet high-quality standards.
The effect is analyzed in~\cref{sec:filtered-data-selection}. 


To mitigate the catastrophic forgetting issue of meta-skill, the meta-skill learning data $D_{\text{meta}}$ are also included in self-evolution training. 
At \( t^{th}\) iteration, the model undergoes self-evolution training with the updated self-curated data $D^{t}_{\text{evol}}$, improving its performance and aligning it more closely with human values. 
\subsubsection{Mathematical Modeling}
\label{sec:self-evolve-training}
\textbf{Main Objective.} 
We denote \( \tau^t_{\phi} \) as the probability distribution generated by \( M_{\text{evol}}^{t} \) with parameters \( \phi \).
The self-evolution training loss \( \mathcal{L}_{\text{evol}}^t(\phi) \) is defined as:
\begin{align}
   \label{eq:self-evolution training}
   \begin{aligned}
\mathcal{L}_{\text{evol}}^t(\phi) &= -\mathbb{E}_{p_{\text{evol}}}\mathbb{E}_{\hat{r}_{\text{evol}} \sim \Psi^{t-1}(\hat{r}_{\text{evol}}|p_{\text{evol}})}\left[\log\tau^t_\phi(\hat{r}_{\text{evol}}|p_{\text{evol}})\right]
\\
&=-\mathbb{E}_{p_{\text{evol}}} \left[ {\sum_{\hat{r}_{{\text{evol}}}} \Psi^{t-1}(\hat{r}_{\text{evol}}|p_{\text{evol}}) \log\tau^t_\phi(\hat{r}_{\text{evol}}|p_{\text{evol}})} \right],
\end{aligned}
\end{align} 
where  \( p_{\text{evol}} \) is sampled from unlabeled prompts corpus (detiled in \cref{app:self-evolution unlabeled prompts}) for self-evolution \(t^{th}\) round. The joint probability distribution is:
\begin{align}
\label{eq:self-refinement-process}
\begin{aligned}
    &\Psi^{t-1}(\hat{r}_{\text{evol}}|p_{\text{evol}})   =
    \\
    &\sum_{r_{\text{evol}}, f_{\text{evol}}} \left(\tau^{t-1}_\phi(r_{\text{evol}}|p_{\text{evol}})  \cdot \tau^{t-1}_\phi(f_{\text{evol}}|r_{\text{evol}}, p_{\text{evol}})   \cdot  \tau^{t-1}_\phi( \hat{r}_{\text{evol}} |f_{\text{evol}}. r_{\text{evol}}, p_{\text{evol}}) \right).
\end{aligned}   
\end{align}
The rationale behind learning from \(\Psi^{t-1}(\hat{r}_{\text{evol}}|p_{\text{evol}})\) is discussed in~\cref{sec:why-design-self-evolution-training}.

Optimizing~\cref{eq:self-evolution training} is equivalent to minimizing the Kullback-Leibler (KL) divergence: 
\begin{align}
\label{eq:kl-divergence-of-self-evolution-models}
\begin{aligned}
&\text{KL}(\Psi^{t-1}(\hat{r}_{\text{evol}}|p_{\text{evol}}) ||\tau^t_\phi(\hat{r}_{\text{evol}}|p_{\text{evol}})) 
\\
&= \sum_{\hat{r}_{{\text{evol}}}} \Psi^{t-1}(\hat{r}_{\text{evol}}|p_{\text{evol}}) \log \frac{\Psi^{t-1}(\hat{r}_{\text{evol}}|p_{\text{evol}})}{\tau^t_{\phi}(\hat{r}_{\text{evol}}|p_{\text{evol}})}
\\
&= - \underbrace{H(\Psi^{t-1}(\hat{r}_{\text{evol}}|p_{\text{evol}}))}_{\text{constant for fixed $\Psi^{t-1}$}} -
\\
&\underbrace{\sum_{ \hat{r}_{{\text{evol}}}} \Psi^{t-1}(\hat{r}_{\text{evol}}|p_{\text{evol}}) \log\tau^t_\phi(\hat{r}_{\text{evol}}|p_{\text{evol}}}_{\text{eq.}~(\ref{eq:self-evolution training})}).
\end{aligned}
\end{align}
The optimization of KL divergence is to fine-tune the model parameters \(\phi\) to ensure that the model's predictive probability distribution \(\tau^t_\phi\) aligns with the joint probability of the preceding iteration's chain process (\(\Psi^{t-1}\)).
The goal is to enhance the model's ability to directly produce refined responses (\(\hat{r}_{\text{evol}}\)) in the \(t^{th}\) iteration, effectively condensing the intricate process of generation, feedback, and refinement from the \((t-1)^{th}\) iteration. 
This advancement demonstrates the model's evolving capability to streamline the complex steps into a more straightforward inference.
The potential plateau is discussed in \cref{sec:Potential limited plateau}.


\textbf{Further Analysis.}
Assuming that each self-evolution round is effective, implying that as \(t\) increases, the quality of responses sampled from \(\Psi^{t}\) improves, optimizing the KL divergence as described in~\cref{eq:kl-divergence-of-self-evolution-models} is fundamentally a process aimed at enhancing the direct generation of high-quality responses.
Before delving deeper, it is crucial to introduce several key concepts.
We define a binary variable \(X\) to evaluate the quality of responses.
A higher probability, \( p(X=1 \mid p_{\text{evol}}, \hat{r}_{\text{evol}}) \), indicates a higher quality of the response \(\hat{r}_{\text{evol}}\) in relation to the prompt \(p_{\text{evol}}\). 
For the self-evolving model with parameters \(\phi\) at the \(t^{th}\) iteration, the model's log-likelihood of producing high-quality responses to a specified prompt is defined as follows:

\begin{align*}
    \begin{aligned}
    \log p^{t}(X=1\mid{p}_{\text{evol}})  :=\log\sum_{\hat{r}}p(X=1\mid p_{\text{evol}},\hat{r}_{\text{evol}})\tau_{\phi}^t(\hat{r}_{\text{evol}}|p_{\text{evol}})
    \end{aligned}.
\end{align*}
By minimizing the KL divergence in~\cref{eq:kl-divergence-of-self-evolution-models}, we can increase \(\log p^{t}(X=1\mid{p_{\text{evol}}})\) by progressively improving its Evidence Lower Bound (ELBO): 
\begin{align*}
\begin{aligned}
&\log p^{t}(X=1\mid{p_{\text{evol}}}) 
\\
&=\log\sum_{\hat{r}_{\text{evol}}}p(X=1\mid p_{\text{evol}},\hat{r}_{\text{evol}})\tau_{\phi}^t(\hat{r}_{\text{evol}}|p_{\text{evol}}).
\\
&=\log\mathbb{E}_{\Psi^{t-1}(\hat{r}_{\text{evol}}|p_{\text{evol}})}\left[\frac{p(X=1\mid p_{\text{evol}},\hat{r}_{\text{evol}})\tau_{\phi}^t(\hat{r}_{\text{evol}}|p_{\text{evol}})}{\Psi^{t-1}(\hat{r}_{\text{evol}}|p_{\text{evol}})}\right]
\\
&\geq\mathbb{E}_{\Psi^{t-1}(\hat{r}_{\text{evol}}|p_{\text{evol}})}\left[\log\frac{p(X=1\mid p_{\text{evol}},\hat{r}_{\text{evol}})\tau_{\phi}^t(\hat{r}_{\text{evol}}|p_{\text{evol}})}{\Psi^{t-1}(\hat{r}_{\text{evol}}|p_{\text{evol}})}\right] 
\\
&= \mathbb{E}_{\Psi^{t-1}(\hat{r}_{\text{evol}}|p_{\text{evol}})} \left[\log p(X=1\mid p_{\text{evol}},\hat{r}_{\text{evol}})\right] 
\\
&\quad -  \underbrace{\text{KL}(\Psi^{t-1}(\hat{r}_{\text{evol}}|p_{\text{evol}}) ||\tau^t_\phi(\hat{r}_{\text{evol}}|p_{\text{evol}}))}_{\text{eq.}~(\ref{eq:kl-divergence-of-self-evolution-models})}.
\end{aligned}
\end{align*}
The entire self-evolution training process can be viewed as a continuous exploration of inherent model capabilities in generation, self-feedback, and self-refinement, ultimately enhancing the model's ability to generate high-quality responses directly.

\textbf{Overall Objective.}~In the iterative self-evolution process, meta-skills, i.e., the ability to self-feedback and self-refinement, is crucial for guiding the evolution process.
We incorporate $D_{\text{meta}}$ into self-evolution training to mitigate the potential catastrophic forgetting of meta-skills: 

\begin{align*}
\begin{aligned}
 \mathcal{L}^{t}_{\text{meta}}(\phi) = -   \mathbb{E}_{(p, r, f, \hat{r})  \sim D_\text{meta}}   \big[
 \log\tau^{t}_\phi(f|p, r)  +
  \log\tau^{t}_\phi(\hat{r}|p, r, f)    \big].
\end{aligned}
\end{align*}

The total objective for the \(t^{th}\) round of self-evolution is:
\begin{align*}
\begin{aligned}
 &\mathcal{L}^{t}_{\text{self}}(\phi) = \mathcal{L}^{t}_{\text{evol}}(\phi) + \mathcal{L}^{t}_{\text{meta}}(\phi).
\end{aligned}
\end{align*}

\subsection{Response Refinement during Inference}
\label{sec:online-refinement}

Equipped with the meta-skills for self-feedback and self-refinement, the model can conduct self-refinement during inference.
Specifically, the model generates an initial response and then refines it using self-refinement, akin to the method described in \cref{sec:meta-skill}. 
Response refinement during inference consistently improves the model's performance as shown in \cref{sec:main-experiment}.

\section{Experiment Settings}
\label{sec:experiment-settings}
\subsection{Evaluation}
\paragraph{Inference Setting.} 
We adopt two inference settings: (1)  \textbf{Direct Response} (default): the model directly answers the question with a Zero-shot Chain of Thought (CoT) methodology~\citep{KojimaGRMI22}, which is the default setting to evaluate the model capability directly; 
(2) \textbf{Self-Refinement}: during inference, the model self-refines its original answer for once, as described in \cref{sec:online-refinement}. 

\textbf{Benchmarks.}
We evaluate on two mathematical benchmarks to show the efficacy of SELF on complex reasoning tasks, and further verify the generalizability of SELF on two general benchmarks. \textbf{GSM8K}~\citep{cobbe2021gsm8k} contains high-quality, linguistically diverse grade school math word problems crafted by expert human writers, which incorporates approximately 7.5K training problems and 1K test problems. 
The performance is measured by accuracy (\%). 
\textbf{SVAMP}~\citep{pateletal2021nlp} is a challenge set for elementary Math Word Problems (MWP). 
It is composed of 1K test samples. 
The evaluation metric is accuracy (\%). 
\textbf{Vicuna testset}~\citep{Vicunatestset} is a benchmark for assessing instruction-following models, containing 80 examples across nine skills in mathematics, reasoning, and coding. 
\textbf{Evol-Instruct testset}~\citep{xu2023wizardlm} includes 218 real-world human instructions from various sources, offering greater size and complexity than the Vicuna testset. 

\subsection{Setup and Baselines}
\label{sec:baselines}
The \textbf{complete SELF framework} includes meta-skill training with $D_{\text{meta}}$, three iterations of self-evolution training, and optional self-refinement during inference. 
Our evaluation primarily focuses on assessing how self-evolution training can progressively enhance the capabilities of LLMs. 
For building the meta-skill training corpus $D_{\text{meta}}$, we employ GPT-4 as the language model labeler \(L\)  due to its proven proficiency in refining responses~\citep{an2023learning} via the prompt described in~\cref{sec:Prompt-feedback-refinement}\footnote{Separate prompts have been designed for the math domain~\cref{sec:Prompt-feedback-refinement-math} and general domain~\cref{app:prompt-general-domain}.}.
The data statistic of \(D_{\text{meta}}\) is shown in \cref{app:dmeta data generation} and further details of unlabeled corpus construction is described in \cref{app:self-evolution unlabeled prompts}.
We note that all model training utilized the same training hyperparameters, as shown in \cref{app:training-hyperparameters}. 

We note that the SELF framework is compatible with versatile LLMs.
In this study, we perform the experiment with \textbf{Vicuna-7b}~\citep{vicuna2023},
a capable open-source instruction-following model fine-tuned from LLaMA-7b~\citep{touvron2023llama}, will be referred to simply as ``Vicuna'' in subsequent sections.
To verify the generalizability of SELF, we also experiment with OpenLLaMA~\cite{openlm2023openllama} and Vicuna-1.5~\citep{vicuna2023} in \cref{app:self-scalability}.
All the compared baselines are outlined:

\textbf{(1) Vicuna + \(D_{\text{QA}}\):} 
To demonstrate the improvement brought by SELF and exclude the impact of standard domain-specific supervised fine-tuning (SFT), we set a direct baseline that trained solely on pseudo-labeled question-answer pairs in the meta-skill training corpus. Specifically, we construct \(D_{\text{QA}}\), which includes all the (\(p, \hat{r}\)) pairs from  \(D_{\text{meta}}\), and fine-tune the model as:
\begin{align*}
\mathcal{L}_{\text{QA}}(\phi) = -\mathbb{E}_{(p, \hat{r}) \sim D_{\text{QA}}} \left[ \log\tau_\phi(\hat{r}|p) \right].
\end{align*}
We refer to this approach as Vicuna + \(D_{\text{QA}}\), the most straightforward baseline. 
The performance gap between  Vicuna + \(D_{\text{QA}}\) and SELF verify the efficacy of the proposed SELF framework, excluding the effect of training on domain-specific QA data.



\textbf{(2) RLHF:} we utilize the RLHF implementation from trlx\footnote{https://github.com/CarperAI/trlx}. 
We apply the same SFT model as the policy model for RLHF, \textbf{Vicuna + \(D_{\text{QA}}\)} as described above, which is consistent with SELF. The reward model is initialized from \textbf{Vicuna-7b} and is fine-tuned using pair-wise comparison data derived from the meta-skill training corpus $D_{\text{meta}}$~(\cref{sec:meta-skill-corpus}), where the refined response $\hat{r}$ is presumed to be better than the original one $r$. 

\textbf{(3) Self-Consistency:}
we compare the self-refinement inference strategy in SELF with the Self-Consistency~\citep{wang2022selfconsistency}~(i.e., multiple sampling and selecting an answer with majority voting) and examine their combined efficacy.


\subsection{Main Result}
\label{sec:main-experiment}

\subsubsection{Math Test}
\begin{table}[ht]
\caption{Experiment results on GSM8K and SVAMP compare SELF with other baseline methods. We evaluate the impact of Self-Evolution (SE), Self-Consistency (SC), and Self-Refinement (SR) strategies on model performance.}
      \label{tab:main-table}
    \vskip 0.15in
    \begin{center}
    \begin{small}
    \begin{adjustbox}{width=0.7\textwidth}
    \begin{tabular}{l c c c c c}
    \toprule
    Model &  SE & SC & SR & GSM8K(\%) & SVAMP(\%)  \\
    \midrule
    \multirow{3}{*}{Vicuna} & & & & 16.43 & 36.40 \\
                            & & \checkmark & & 19.56 & 40.20 \\
                            & & & \checkmark & 15.63 & 36.80 \\
    \midrule
    \multirow{3}{*}{Vicuna + \(D_{\text{QA}}\)} &  & & & 24.49 & 44.90 \\
                                         &  & \checkmark & & 25.70 & 46.00 \\
                                         &  & & \checkmark & 24.44 & 45.30 \\
    \midrule
    \multirow{3}{*}{Vicuna + SELF (Ours)} & \checkmark & & & 29.64 & 49.40 \\
                                         & \checkmark & \checkmark & & 29.87 & 50.20 \\
                                         & \checkmark & & \checkmark & 31.31 & 49.80 \\
                                         & \checkmark & \checkmark & \checkmark & \textbf{32.22} & \textbf{51.20} \\
                                         
    \bottomrule
    \end{tabular} 
    \end{adjustbox}
  
     \end{small}
   \end{center}
\end{table}

In \cref{tab:main-table}, we compare SELF against baseline models, as detailed in \cref{sec:baselines}. 
This comparison elucidates SELF's effectiveness in enhancing LLM performance through self-evolution and offers several key insights:

\textbf{(1) Self-Evolution Enhances LLM:} Vicuna + SELF significantly outperforms its baseline  Vicuna + \(D_{\text{QA}}\) ($24.49\%\xrightarrow{+5.15\%}29.64\%$ on GSM8K and $44.90\%\xrightarrow{+4.5\%}49.40\%$ on SVAMP) in direct response setting, showcasing self-evolution is effective in optimizing LLMs.

\textbf{(2) SELF Instills Self-Refine Capability in LLMs:} The integration of self-refinement inference strategy with Vicuna  + SELF further boosts performance ($29.64\%\xrightarrow{+1.67\%}31.31\%$), while baseline models show marginal or negative effect via self-refinement.
We also provide a case analysis for the limited self-refinement ability of baseline models, as shown in~\cref{Fig:self-case-analysis}.
This indicates that SELF can instill advanced self-refinement capabilities into small LLMs like Vicuna (7B), although self-refinement was previously shown as an exclusive ability of large LLMs~\citep{selfee2023} like GPT-4.

\textbf{(3) SELF can work with Self-Consistency:} 
SELF works effectively with self-consistency, improving accuracy across models.
The base Vicuna model, which may have uncertainties in its outputs, shows notable improvement with self-consistency, achieving a +3.13\% increase.
As the model progresses through self-evolution training and becomes more certain of generating correct math answers, the benefit from self-consistency reduces.
Combining self-refinement with self-consistency further elevates performance (e.g., $29.64\%\xrightarrow{+2.58\%}32.22\%$ on GSM8K), indicating that these two strategies can complement each other effectively.

\textbf{(4) Pseudo-Labeled $D_{\text{QA}}$ Enhances Performance:} The inclusion of pseudo-labeled QA data $D_{\text{QA}}$ enhances Vicuna's performance, suggesting that tuning with domain-specific QA data can enhance task-specific problem-solving.

\subsubsection{Comparison with RLHF}
\label{sec:exp-compare-rlhf}
\begin{table}[h]
    \caption{Comparison of SELF and RLHF on GSM8K. ``Feedback Acc.'' measures how accurately feedback identifies correct and incorrect answers, while ``GSM8K Acc.'' shows the model performance on GSM8K testset.}
    \label{tab:self-RLHF}
    \vskip 0.15in
    \begin{center}
    \begin{small}
    \begin{adjustbox}{width=0.6\textwidth}
    \begin{tabular}{l c c}
        \toprule
        Method & {Feedback Acc.(\%)} & {GSM8K Acc.(\%)}   \\
        \midrule
        Vicuna + $D_{\text{QA}}$ & - & 24.49  \\
        RLHF &  24 & 25.55  \\
        SELF &  \textbf{72} & \textbf{27.67}  \\
        \bottomrule
    \end{tabular}
    \end{adjustbox}
    \end{small}
   \end{center}
\end{table}

In \cref{tab:self-RLHF}, we compare the performance of SELF with RLHF. 
To alleviate the effect led by different amounts of training data and make a fair comparison, for SELF, we only adopt data solely from the initial round of self-evolution training. This ensures the same training data quantity with RLHF and leads to sub-optimal results compared with the one in \cref{tab:self-RLHF}.
As \cref{tab:self-RLHF} shows, RLHF achieves a 25.55\% accuracy on GSM8K, which is lower than the 27.67\% performed by SELF.
We observe that the simple scalar reward of RLHF often fails to identify the correctness of the reasoning process, which limits performance improvements. 
On the GSM8K test set, for incorrect answers produced by the SFT model (Vicuna + $D_{\text{QA}}$), the reward model only identifies 24\% of them as incorrect, i.e., the reward model assigns lower scalar rewards to incorrect answers compared to correct answers.
In contrast, SELF utilizes informative natural language feedback to provide a more accurate assessment. It correctly identifies 72\% of incorrect answers.
 
\subsection{General Test}
We test the efficacy and generalizability of SELF on general domain benchmarks, explicitly using the Vicuna and Evol-Instruct test sets. 
Three configurations of the Vicuna model are evaluated: Vicuna, Vicuna + $D_{\text{QA}}$, and Vicuna + SELF.
We utilize GPT-4 to evaluate the models' responses on both test sets. We follow the assessment methodology proposed by~\citep{xu2023wizardlm}, which mitigated the order bias presented in the evaluation procedures. 

The results are depicted in Figure \ref{fig:vicuna-and-evol-instruct-testset}. In the figure, \textcolor{blue}{blue} represents the number of test cases where the model being evaluated is preferred over the baseline model (Vicuna), as assessed by GPT-4. 
\textcolor{yellow}{Yellow} denotes test cases where both models perform equally, and \textcolor{pink}{pink} indicates the number of test cases where the baseline model is favored over the model being evaluated.

    

\begin{figure}[ht]
    \centering
    \begin{subfigure}[b]{0.495\textwidth}
        \includegraphics[width=\textwidth]{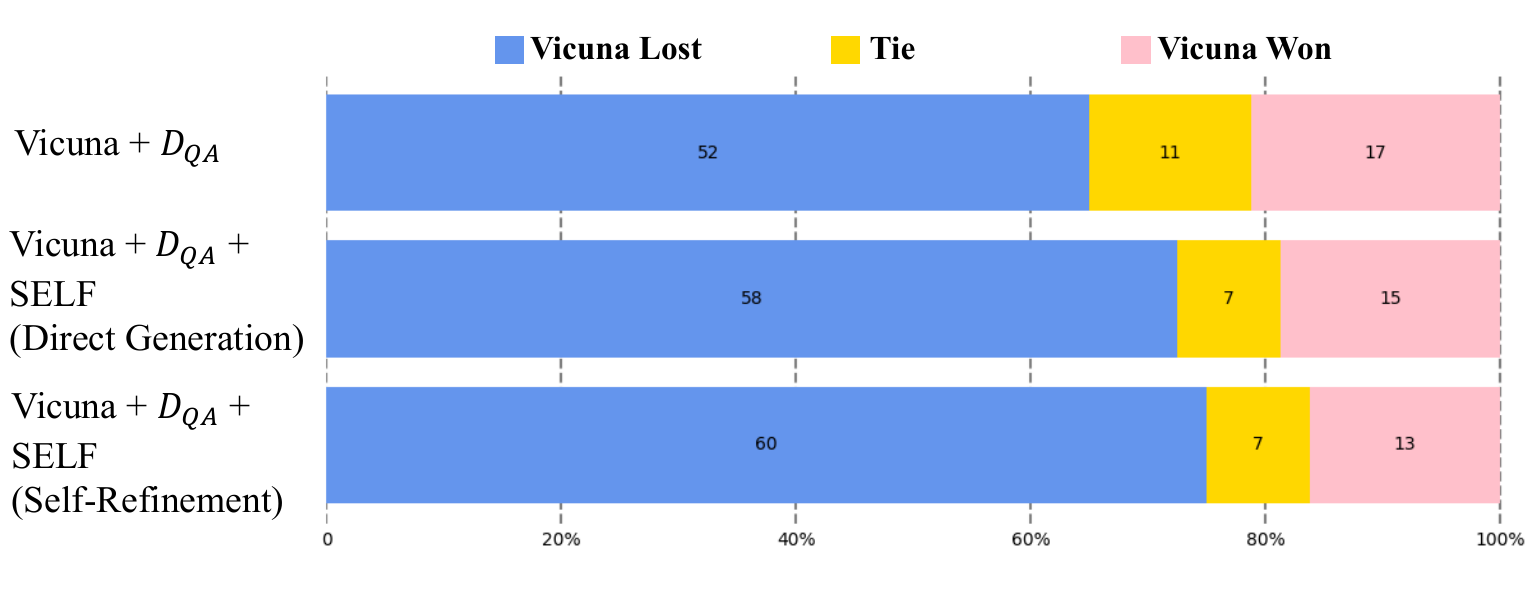}
        \caption{Results on Vicuna testset.}
        \label{fig:vicuna-testset}
    \end{subfigure}
    \hfill 
    \begin{subfigure}[b]{0.495\textwidth}
        \includegraphics[width=\textwidth]{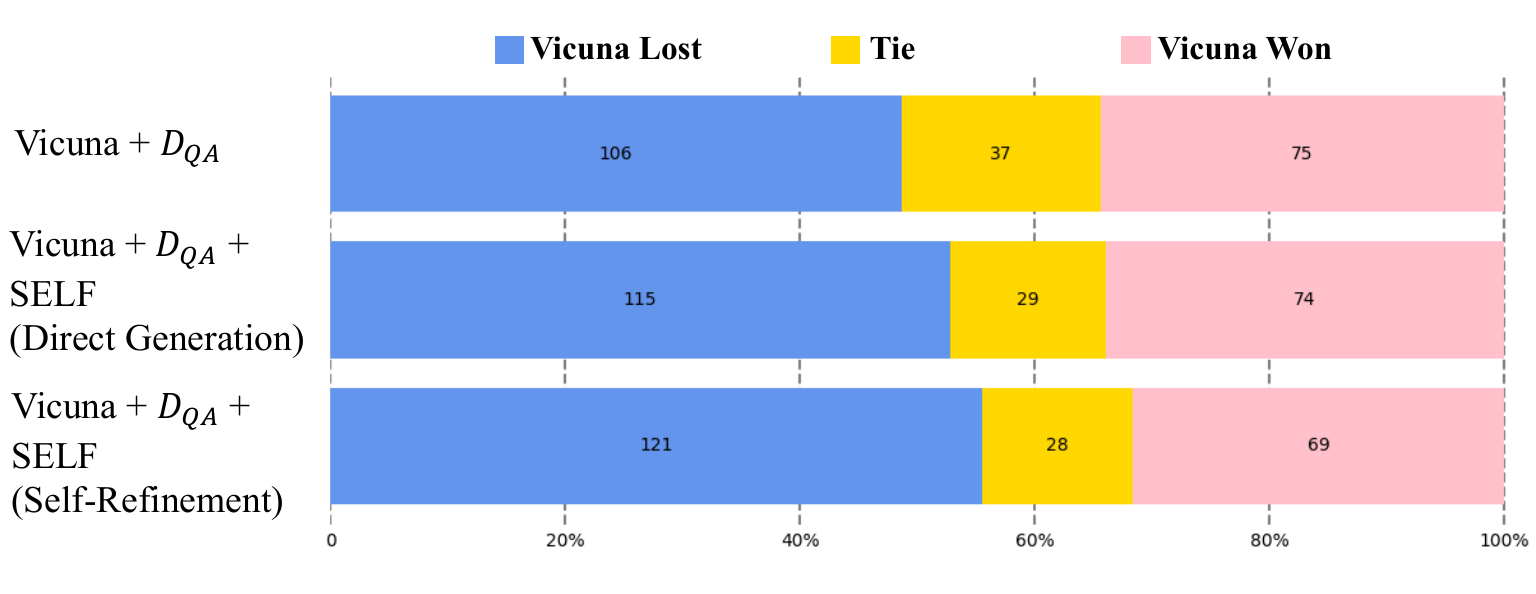}
        \caption{Results on Evol-Instruct testset.}
        \label{fig:evol-instruct-testset}
    \end{subfigure}
\caption{Results on Vicuna testset and Evol-Instruct testset}
    \label{fig:vicuna-and-evol-instruct-testset}
\end{figure}

In the Vicuna testset, SELF increases direct response win rate from 65.0\% to 72.5\% compared with Vicuna + $D_{\text{QA}}$. 
After self-refinement, the win rate is further improved to 75.0\%.  In the Evol-Instruct testset, the win rate of Vicuna + $D_{\text{QA}}$ is 48.6\%. 
SELF increases the win rate to approximately 52.8\%. 
Applying self-refinement during inference further improves the win rate to 55.5\%.

These findings in the general domain highlight the SELF framework's adaptability and robustness, particularly when self-refinement is employed, showcasing its efficacy across varied test domains.




\subsection{Ablation Study}
\label{sec:self-evolution-experiment}
\begin{table}[ht]
\caption{Performance under various training settings of SELF. A checkmark $\checkmark$ in a column denotes the additive adoption of the corresponding setting in that training scenario. We present two  kinds of inference results: \textbf{Direct Response}~(DR) and \textbf{Self-Refinement}~(SR), the latter conducts self-refinement to DR.}

\label{tab:self-evolve}
    \begin{center}
    \begin{small}
     \begin{adjustbox}{width=0.7\textwidth}
    \begin{tabular}{ccccccccc}
        \toprule
        \multicolumn{2}{c}{SVAMP (\%)} & \multicolumn{2}{c}{GSM8K (\%)} &   \multirow{2}{*}{ $D_{\text{QA}}$} &  \multirow{2}{*}{ $D_{\text{meta}}$} & \multicolumn{3}{c}{Self Evol.} \\
         \cmidrule(lr){1-2}   \cmidrule(lr){3-4} \cmidrule(lr){7-9}
       DR& SR& DR & SR &  &  & 1st  & 2nd   & 3rd \\
        \midrule
        \midrule
       36.4 & 36.8 & 16.43 & 15.63 & & & & &  \\
       44.9  &  45.3 & 24.49 & 24.44 &$\checkmark$  & & & &  \\
       46.8 & 47.0 & 25.39 & 28.28 & & $\checkmark$ & & & \\
       47.8 & 48.0 & 27.67 & 29.34 & & $\checkmark$ & $\checkmark$ & & \\
       48.9 & 49.0 & 28.66 & 29.87 & & $\checkmark$ & $\checkmark$ & $\checkmark$ & \\
       \textbf{49.4} & \textbf{50.2} & \textbf{29.64} & \textbf{31.31} & & $\checkmark$ & $\checkmark$ & $\checkmark$ & $\checkmark$ \\

        \bottomrule
    \end{tabular} 
    \end{adjustbox}
     \end{small}
   \end{center}
\end{table} 
We conduct ablation experiments on SVAMP and GSM8K datasets to assess the incremental effect of each stage. 
While baseline models exhibit slight or even adverse effects via self-refinement, the SELF framework endows LLMs with an inherent capability through meta-skill learning and multi-iterations of self-evolution training. 
As depicted in \cref{tab:self-evolve}, our framework facilitates gradual performance improvements through successive SELF stages. Observations are highlighted below:

\textbf{(1) Meta-skill Training Elevates Performance:} Training with the meta-skills dataset \(D_{\text{meta}}\) as defined in~\cref{eq:meta-skill learning}, and setting \(\beta = 1\) for a fair comparison with the question-answer dataset \(D_{\text{QA}}\), improves \textbf{direct response} performance. Specifically, we observe an increase of +0.90\% on the GSM8K dataset and +1.9\% on the SVAMP dataset, compared to using the \(D_{\text{QA}}\) dataset alone. 
This underscores the interesting finding that arming the model with self-feedback and self-refinement meta-capability implicitly elevates its capacity to generate superior responses directly, even without explicit self-refinement. 
We offer an insight in~\cref{sec:why-Meta-skill-Training-helps-qa-performance}.

\textbf{(2) Continuous Improvement through Self-Evolution:} The results reveal that three self-evolution rounds consecutively yield performance enhancements (e.g., $25.39\%\xrightarrow{+2.28\%}27.67\%\xrightarrow{+0.99\%}28.66\%\xrightarrow{+0.98\%}29.64\%$ on GSM8K). This shows that the model actively self-evolves, refining its performance autonomously without additional manual intervention.

\textbf{(3) Persistent Efficacy of Self-Refinement:} After meta-skill learning, regardless of model variation, executing self-refinement consistently results in notable performance improvements. This shows that the self-refinement meta-capability learned by SELF is robust and consistent across evolution steps.

\subsection{Analysis on Data Filtering with Self-Feedback}\label{sec:filtered-data-selection}
\begin{table}[h]
\small
\caption{Impact of Data Filtering with Self-Feedback on GSM8K. ``Training Acc.'' shows the accuracy of the self-evolution training data post-filtering, and ``Test Acc.'' represents the model's test performance post-training on these filtered data.}
 \vskip 0.15in
\centering
    \begin{center}
    \begin{small}
\begin{adjustbox}{width=0.55\textwidth}{

\begin{tabular}{lcc}
\toprule
{Filter Strategy} & {Training  Acc. (\%)} & {Test Acc. (\%)} \\
\midrule
Unfiltered & 29.89 & 26.90 \\
Filtered & \textbf{44.10} & \textbf{27.67}  \\
\bottomrule
\end{tabular}
}
\end{adjustbox}
    \end{small}
    \end{center}

\label{tab:expanded_filtering_analysis}
\vspace{-0.1in}
\end{table}
Table~\ref{tab:expanded_filtering_analysis} presents an analysis of filtering self-evolution training data using self-feedback~(\cref{sec:self-evolve-training-corpus}) on GSM8K, focusing on training data quality and its influence on self-evolution training. 
The filtering criteria are detailed in~\cref{app:qualified function design}.

The following insight is highlighted: The combination of self-refinement and self-feedback filtering results in higher self-evolution training data accuracy (+14.21\%) and improved fine-tuned model performance (+0.77\%). Despite the significant training data accuracy improvement, the performance gain is modest due to the reduced data size (from 4K to 1.8K) after filtering.


\section{Conclusion}
We present SELF (Self-Evolution with Language Feedback), a novel framework that enables LLMs to achieve progressive self-evolution through self-feedback and self-refinement.
Unlike conventional methods, SELF transforms LLMs from passive information recipients to active participants in their evolution.
The adoption of natural language feedback promotes a more informative and fine-grained evaluation.
Through meta-skill learning, SELF equips LLMs with the capability for self-feedback and self-refinement. 
This empowers the models to evolve their capabilities autonomously, utilizing self-evolution training and online self-refinement.
Experiments conducted on benchmarks underscore SELF’s capacity to progressively enhance model capabilities while reducing the need for human intervention.
SELF represents a leading step in autonomous LLM development, leading to an insight that models are capable of continual learning and self-evolution.

\bibliography{main}
\bibliographystyle{Paper}


\appendix

\section{Appendix}

\subsection{Discussion}

\subsubsection{Why Refinement is Better}
\label{sec:why-design-self-evolution-training}
We discuss why it's necessary to optimize \(\tau^{t}_{\phi}(\hat{r}_{\text{evol}}|p_{\text{evol}})\) in the \(t^{th}\) round self-evolution by learning from \(\Psi^{t-1}(\hat{r}_{\text{evol}}|p_{\text{evol}})\), and why we believe samples from \(\Psi^{t-1}(\hat{r}_{\text{evol}}|p_{\text{evol}})\) are typically of higher quality than those from \(\tau^{t-1}_{\phi}({r}_{\text{evol}}|p_{\text{evol}})\) directly.

Firstly, similar to the insights analyzed in \cref{sec:training-objective}, we believe that a process akin to CoT, involving feedback followed by refinement before providing an answer, helps in generating high-quality responses. Secondly, \(r_{\text{evol}}\) is already a reasonably good output after meta-skill learning and previously (\(t-1\)) rounds of self-evolution. We can assume that the self-feedback \(f_{\text{evol}}\) is informative, hence \(\hat{r}_{\text{evol}} \sim \tau^{t-1}_\phi(\hat{r}_{\text{evol}} | p_{\text{evol}}, r_{\text{evol}}, f_{\text{evol}})\) is of higher quality than \(r_{\text{evol}} \sim \tau^{t-1}_{\phi}(r_{\text{evol}}|p_{\text{evol}})\) because it incorporates useful feedback information. 
If \(f_{\text{evol}}\) suggests that the initial response \(r_{\text{evol}}\) does not require refinement, we still proceed through the process of revising from \(r_{\text{evol}}\) to \(\hat{r}_{\text{evol}}\) using \(f_{\text{evol}}\), but set \(\hat{r}_{\text{evol}} = r_{\text{evol}}\). By doing so, we ensure that the quality of \(\hat{r}_{\text{evol}}\) is at least as good as that of \(r_{\text{evol}}\).

Moreover, as described in \cref{sec:self-evolve-training}, we utilize \textbf{Data Filtering with Self-feedback}. In other words, we only keep \(\hat{r}_{\text{evol}}\) evaluated as \textit{qualified}, allowing us to emphasize high-quality outputs and further improve \(\tau_{\phi}^{t} \).

\subsubsection{Why Integration of Meta-skill Training Data \texorpdfstring{\(D_{\text{meta}}\)} Elevates Direct QA}
\label{sec:why-Meta-skill-Training-helps-qa-performance}
The \(D_{\text{meta}}\) dataset trains the model to not only modify answers but also to fully grasp a prompt, create feedback, and then develop a revised answer. This approach resembles training the model to think through a problem in a chain-of-thought methodically (CoT) manner, before responding. The training encompasses a thorough examination of the entire process, which not only betters the model's direct response capability but also enriches its understanding of the logic behind those answers, thereby enhancing its generalization ability.


\subsubsection{Potentially Limited Plateau of Self-evolution Training}
\label{sec:Potential limited plateau}
Based on \cref{eq:self-evolution training} and \cref{eq:self-refinement-process}, the model in the \( t^{th} \) round is updated to improve direct response quality by incorporating the generate-feedback-refinement process from the \( (t-1)^{th} \) round. This is based on the assumption that the refined response is superior to the initial one generated by \( M_{\text{evol}}^{t-1} \). As illustrated in Fig.~\ref{Fig:main_picture}, the direct generation performance of \( M_{\text{evol}}^{t} \) (green curve) consistently falls below the self-refinement of \( M_{\text{evol}}^{t-1} \) (blue curve). The self-refinement gains in the \( (t-1)^{th} \) round indicate the potential benefit that the \( t^{th} \) round self-evolution could bring to direct generation. This also helps determine when to halt the self-evolution process, i.e., the process can be stopped when self-refinement brings no benefit to the direct response.

\subsection{Prompt of Generating Feedback and Refinement for Meta-skill Corpus}
\label{sec:Prompt-feedback-refinement}
We introduce the prompt for generating feedback and refinement in two domains: Math and General. We outline specific prompts designed to guide the evaluation and improvement of responses to questions for building \(D_{\text{meta}}\) in each domain.
\subsubsection{Math Domain}
\label{sec:Prompt-feedback-refinement-math}
For the Math Domain, the prompt instructs evaluators to assess the quality of a response to a math question, provide a step-by-step analysis, and determine its correctness. If the response is incorrect, the evaluator is asked to refine and provide a correct answer.
\begin{tcolorbox}
 \label{Fig:feedbackandreviseprompt}
\small
    \textbf{Prompt for feedback and refinement:}
    
    \textbf{(Feedback)} Please assess the quality of the response to the given question. 
    
    Here is the question: {$p$}. 
    
    Here is the response: {$r$}.
    
    Firstly, provide a step-by-step analysis and verification for response starting with ``Response Analysis:''. 
    
    Next, judge whether the response correctly answers the question in the format of ``judgment: correct/incorrect''.
    
    \textbf{(Refinement)} If the answer is correct, output it. Otherwise, output a refined answer based on the given response and your assessment.

\end{tcolorbox}
\subsubsection{General Domain}
\label{app:prompt-general-domain}
For the general test, aligned with the methodology described in~\cref{sec:method}, we deploy the following prompt to guide an LLM-based annotator in generating response feedback and refinement. This prompt serves as the foundation for the meta-skill learning corpus and assists in producing self-evolution training data in the general test setting.
\begin{tcolorbox}
\small
    \textbf{Prompt for feedback and refinement:}
    
    \textbf{(Feedback)} Please assess the quality of response to the given question. 
    
    Here is the question: {$p$}. 
    
    Here is the response: {$r$}.
    
    Firstly provide an analysis and verification for response starting with ``Response Analysis:''. 
    
    Next, then rate the response on a scale of 1 to 10 (1 is worst, 10 is best) in the format of ``Rating:"
    
    \textbf{(Refinement)} Finally output an improved answer based on your analysis if no response is rated 10.
    \label{Fig:feedbackandrevisepromptgeneral_general}
\end{tcolorbox}

\subsection{Data Generation} 

\subsubsection{ \texorpdfstring{\(D_{\text{meta}}\)} Data Quantity}
\label{app:dmeta data generation}
The \(D_{\text{meta}}\) dataset was generated using 3.5k unlabeled prompts from GSM8K and 2K from SVAMP\footnote{Adhering to the official recommendation~\url{https://github.com/arkilpatel/SVAMP/tree/main}, training prompts consist of MAWPS~\citep{koncel2016mawps} and ASDiv-A~\citep{miao2020diverse}}. 

For general tests, 6K conversations were selected from 90K ShareGPT dialogues to form the general \(D_{\text{meta}}\) data.

\subsubsection{Unlabeled Prompts for Self-Evolution Training}
\label{app:self-evolution unlabeled prompts}
\textbf{Math Domain:}
For math tests, unlabeled prompts in self-evolution training were sourced as follows:

(1)~{First round self-evolving phase:} 4K leftover prompts from GSM8k and 1K from SVAMP, excluding those used in \(D_{\text{meta}}\).

(2)~{Second/Third rounds:} 10K/15K prompts were generated using Self-Instruct method~ \citep{wang2022self}, based on a template shown in \cref{Fig:self-instruct-prompt} with initial 4 to 6 seed examples.

\textbf{General Domain:} 15K unlabeled prompts from ShareGPT dialogues were used for self-evolution training data construction.

\begin{tcolorbox}
\label{Fig:self-instruct-prompt}
  You are an experienced instruction creator. You are asked to develop 3 diverse instructions according to the given examples.

  Here are the requirements:
  
  1. The generated instructions should follow the task type in the given examples.
  
  2. The language used for the generated instructions should be diverse.

  Given examples: \{examples\}

  The generated instructions should be:
  
  A. ...
  
  B. ...
  
  C. ...
\end{tcolorbox}

\subsection{Training Hyperparameters}
\label{app:training-hyperparameters}

Our experiments were conducted in a computing environment with 8 NVIDIA V100 GPUs, each having 32GB of memory. All models were fine-tuned in a full-parameter setting. We utilized the AdamW optimizer for model training over 3 epochs, with a batch size of 128. The learning rate was set at 2e-5, including a 3\% learning rate warmup period. Below we provide a comprehensive overview of the training hyperparameters employed in~\cref{tab:training-parameters-row}. These parameters were uniformly applied across all training methods in our experiments.

\begin{table}[h]
\caption{Training hyperparameters.}
\vskip 0.15in
    \begin{center}
    \begin{small}
\centering
\begin{tabular}{ccccccc}
\toprule
\textbf{Hyperparameter} & Global Batch Size & LR & Epo. & Max Length & Weight Decay & Warmup Ratio \\
\midrule
\textbf{Value} & 128 & \(2 \times 10^{-5}\) & 3 & 2048 & 0 & 0.03 \\
\bottomrule
\end{tabular}
    \end{small}
    \end{center}
\label{tab:training-parameters-row}
\end{table}

\subsection{Case Study Analysis}
\label{sec:self-case-analysis} 
\begin{figure}[ht] 
\centering 
\includegraphics[width=0.95\textwidth]{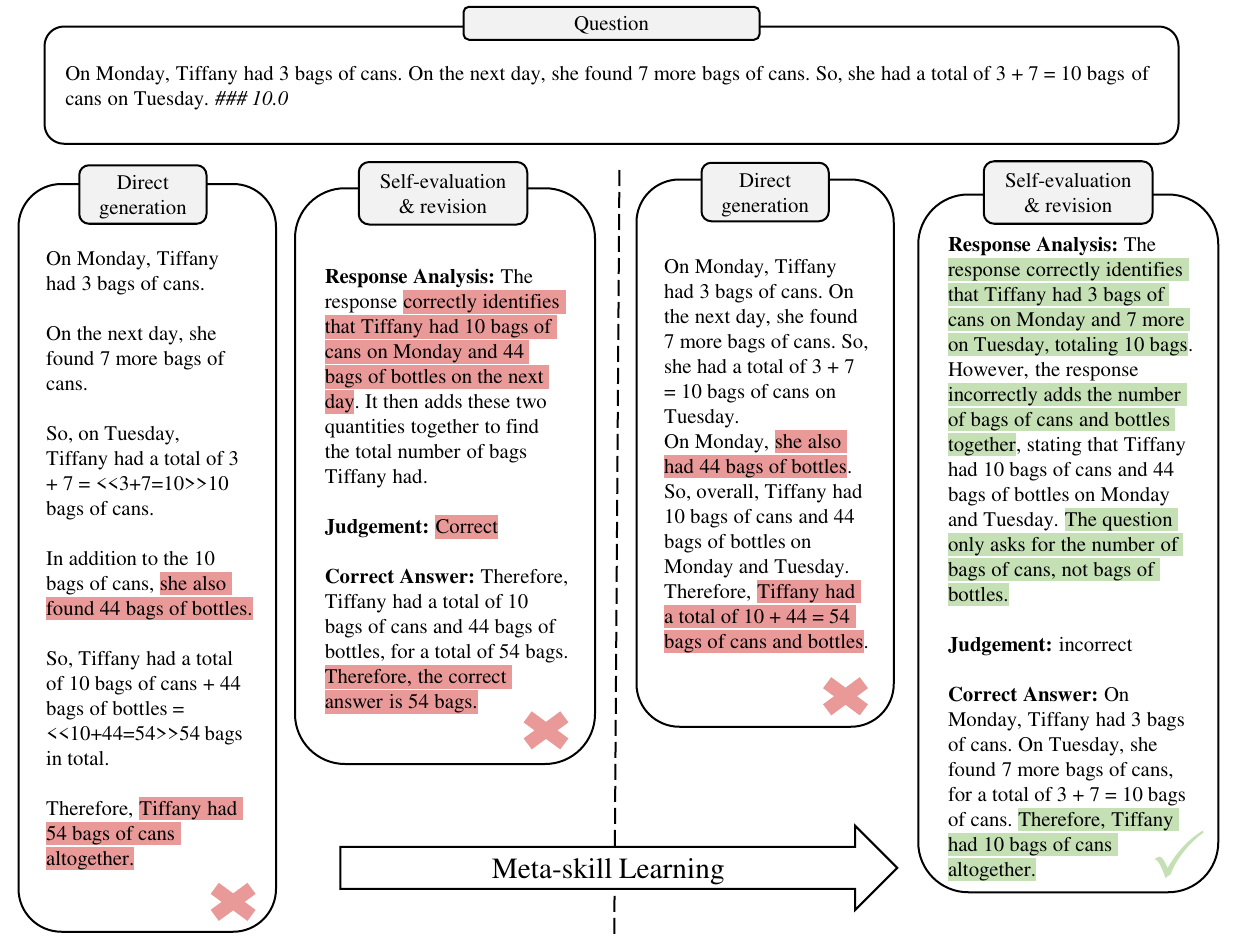}
\caption{Case study comparing the original Vicuna (left) and Vicuna+SELF (right) on a SVAMP problem. Both models generate direct predictions and undergo self-feedback and self-refinement. Both models initially produce answers, followed by self-feedback and self-refinement. Vicuna maintains the incorrect response after refinement, whereas Vicuna+SELF demonstrates enhanced self-refinement, leading to a correct and logically consistent solution.\label{Fig:self-case-analysis}}
\end{figure}
This subsubsection provides an in-depth case study that contrasts the performance of the original Vicuna and Vicuna + SELF models. 
Illustrated in~\cref{Fig:self-case-analysis}, both models perform initial predictions, followed by self-feedback and refinement steps. Notably, Vicuna’s refinement fails to correct its initial errors, while Vicuna + SELF effectively utilizes self-feedback and refinement to derive an accurate and logically coherent answer.

\subsection{Meta-Skill Training Corpus}
\label{app:meta-skill training corpus}
The example shown below exemplifies a standard training example from our meta-skill corpus. It illustrates the model's initial response, followed by its self-feedback, and the ensuing refinement. This process demonstrates how the model is trained for self-feedback and self-refinement capabilities.
\begin{tcolorbox}
\small
    \textbf{Question:} Greg has an alarm set to ring three times a day as a reminder. When the alarm goes off, it continues to ring until Greg turns it off. The first time it went off today, it rang four times. The second time it went off, it rang for three times as long as the first time. The third time, it rang for half as long as the second time. How many times did the alarm ring in all?

    \textbf{Original Response: }The alarm rang four times the first time, three times as long as the first time the second time, and half as long as the second time the third time. So, the total number of times the alarm rang is 4 + 3*3 + 2*2 = 24.

    \textbf{Feedback: } The response correctly interprets the problem. The first time the alarm rang, it rang for 4 times. The second time it rang, it rang for 3 times as long as the first time, which is 3*4 = 12 times. The third time it rang, it rang for half as long as the second time, which is 12/2 = 6 times. However, the final calculation is incorrect. The total number of times the alarm rang is not 4 + 3*3 + 2*2 = 24, but 4 + 12 + 6 = 22. \textit{Judgement: incorrect}

    \textbf{Refined Response:} The alarm rang four times the first time, three times as long as the first time the second time, and half as long as the second time the third time. So, the total number of times the alarm rang is 4 + 12 + 6 = 22.

\end{tcolorbox}

\subsection{Algorithm}
\label{sec:algorithm}
The ``Two-Phase SELF Process'' algorithm outlines a method for developing a base language model through a two-staged approach: Meta-Skill Learning and Self-Evolving. 
The process starts with training on a ``Meta-Skill Learning corpus'', which consists of data representing the generation, feedback and refinement process.
Following this, the model enters the ``Self-Evolving Phase'', where it undergoes iterative refinements, employing data augmentation in each iteration to produce self-refined outputs from its previously refined versions. This iterative self-evolution aims to leverage accumulated knowledge and further enhance the model with newly generated data. The final outcome is an advanced Language Model that has significantly evolved from its original state through multiple self-evolution stages.
More details are delineated in Alg. \ref{alg:self-evolution-process}.
\begin{algorithm}[ht]
\caption{Two-Phase SELF Process}
\small
\label{alg:self-evolution-process}
  \KwData{ (1) Meta-Skill training data (\(D_{\text{meta}}\)) and (2) unlabeled prompts }
  \textbf{Input}: An initial Language Model \(M_{\text{init}}\)
  
  \KwResult{A stronger Language Model \(M^T_{\text{evol}}\) after self-evolving}

  \BlankLine
  \tcp{Meta-Skill Learning Phase}
  \KwData{Meta-Skill learning corpus (\(D_{\text{meta}}\))}
  \(M_{\text{meta}}\) = \text{Supervised\_fine\_tuning}(\(M_{\text{init}}\), \(D_{\text{meta}} \) )\;
  \BlankLine
  
  \tcp{Self-Evolving Phase}
  Initialize \(M^{0}_{\text{evol}}\) with \(M_{{meta}}\)\;
  
  \ForEach{iteration \(t\) in 1 to Number of self-evolving iterations T}{
    \tcp{Data-Augmentation}
    Initialize \(D^{t}_{\text{evol}}\) as an empty set\;
    
    \ForEach{prompt \(p^i_{\text{evol}}\) in \({t}^{th}\) unlabeled prompts}{
      Generate direct output \({r}^i_{\text{evol}}\) using \(M^{t-1}_{\text{evol}}\)\;
      
      Generate self-refined output \(\hat{r}^i_{\text{evol}}\) from \({r}^i_{\text{evol}}\) using \(M^{t-1}_{\text{evol}}\)\;
      
      Use \(M^{t-1}_{\text{evol}}\) to filter the self-refined output\;
      
      Add \((p_{\text{evol}}^i, \hat{r}^i_{\text{evol}})\) to \(D^{t}_{\text{evol}}\), where \(r_i\) is the refined response\;
    }
    \tcp{Self-Evolution Training}
    \(M^{t}_{\text{evol}}\) = \text{Supervised\_fine\_tuning}(\(M^{t-1}_{\text{evol}}\), \(D^{t}_{\text{evol}}\))\;

  }
  \BlankLine
  \tcp{Training Complete}
  \Return Improved Language Model \(M^{T}_{\text{evol}}\)\;
\end{algorithm}

\subsection{Data Filtering Standards}
\label{app:qualified function design}
We design a boolean function, \(\textit{qualified}(f)\), to evaluate feedback \(f\) across different domains, determining if a response to a specific prompt satisfies essential quality criteria.

In the \textbf{Math Domain},the function assesses feedback based on the explicit statement of ``correctness'' in the evaluator's judgment, aligned with the prompt structure in~\cref{sec:Prompt-feedback-refinement-math}. 
It checks if the word ``correct'' immediately follows the phrase ``judgment:'' in the feedback. A presence of ``correct'' results in \(\textit{qualified}(f)\) returning 1, meeting the qualification criteria. Absence leads to a return of 0.

For the \textbf{General Domain}, following the structure in~\cref{app:prompt-general-domain},  \(\textit{qualified}(f)\) extracts and evaluates a numerical rating from the feedback. If the rating, found after "Rating:", is 7 or higher, the function returns 1, indicating qualification. Ratings below 7 return 0, failing to meet the threshold. A rating of 7 balances quality and training data quantity.

\(\textit{qualified}(f)\) is key in both domains for filtering and assessing feedback quality, ensuring only high-quality responses are used for refined answer generation in self-evolution training. 
Post data filtering, \(\Psi^{t-1}\) in~\cref{eq:self-refinement-process} requires an update to \(\Psi'^{t-1} = \Psi^{t-1} \times \textit{qualified}(f)\), adding a quality filter through self-feedback. 
For clarity, we continue using original formulation 
as stated in~\cref{eq:self-refinement-process} in the main text.

\subsection{Multiple v.s. Single Self-Refinement}
\label{sec:single-vs-multiple-revisions}
This study explores the effects of two meta-skill training data organization strategies on model performance: (1) Multiple Self-Refinement ($D_{\text{meta-multi}}$), involving the sampling of three responses for the model to choose the best for refinement, and (2) Single Self-Refinement ($D_{\text{meta}}$), where the model generates and refines a single response.

\cref{tab:comparison-of-single-and-comp-feedback} compares these methods' performances. Both strategies show performance gains with increased training data volume. However, as data volume expands, the multiple-response refinement shows a smaller improvement in direct generation performance (\(+4.02\%\)) than the single-response method (\(+5.84\%\)).
Considering the simplicity and computational efficiency of the single-response method, which only samples one response during inference, and its better performance than the multiple-response approach, we have opted for the single-response refinement strategy in our experiments.

\begin{table}[ht]
\caption{Performance comparison of single and multiple response refinement with varying volumes of meta-skill training data. The arrow indicates improvement from direct generation to self-refinement: ``direct generation $\rightarrow$ self-refinement''.}
\vskip 0.15in
    \begin{center}
    \begin{small}
\centering
\begin{tabular}{lcc}
\toprule
{Data Size} &  {Vicuna + $D_{\text{meta}}$}   &   {Vicuna +  $D_{\text{meta-multi}}$} \\
\midrule
 3.5k & 25.39 $\to$ 28.28 & 25.92 $\to$ 27.29 \\
 7.5k & 31.23 $\to$ 32.98 & 29.94 $\to$ 32.14  \\
\bottomrule
\end{tabular}
    \end{small}
    \end{center}
\label{tab:comparison-of-single-and-comp-feedback}
\end{table}

\subsection{Self-Evolution Training: Continual Training v.s. Restart Training}
\label{app:restart-vs-continual}

\begin{table}[ht]
    \small
    \caption{Analysis about varied self-evolution training methodologies on GSM8K.}
\vskip 0.15in
    \begin{center}
    \begin{small}
    \centering
    \begin{tabular}{lcc}
        \toprule
       {Training Approach} & {Direct Generation (\%)} & {Self-Refinement (\%)} \\
        \midrule
        Base Model &  24.49 & 24.49 \\
        Restart Training & \textbf{27.67} & \textbf{29.34} \\
        Continual Training (Mixed Data) & 27.22 & 28.43 \\
        Continual Training ($D^{t}_{\text{evol}}$ Only) & 24.87 & 25.85 \\
        \bottomrule
    \end{tabular}
    \end{small}
    \end{center}
    \label{tab:training_approaches}
\end{table}
``Restart Training'', which combines meta-skill learning corpus with all rounds of self-evolution training data, significantly improves direct generation (+3.18\%) and self-refinement (+3.85\%).

``Continual Training (Mixed Data)'', where the model is trained simultaneously with all rounds of self-evolution data, also shows notable enhancements in direct generation (+2.73\%) and self-refinement (+3.94\%).
In contrast, ``Continual Training ($D^{t}_{\text{evol}}$ Only)'', which trains the model sequentially with self-evolution data from each round, demonstrates more modest gains (+0.38\% in direct generation, +0.98\% in self-refinement). The relatively lower performance of the latter approach highlights the importance of a mixed data strategy for effective self-evolution training.

Throughout our main text, we have consistently employed the ``Restart Training'' method. This approach was selected for its superior performance, as evidenced in~\cref{tab:training_approaches}. In addition, the integration of \(D_{\text{meta}}\) into the self-evolution training is crucial to prevent the potential catastrophic forgetting of meta-skills. This strategy is essential for preserving the effectiveness and reliability of the self-evolution training process, as highlighted in~\cref{sec:self-evolve-training}.

\subsection{SELF vs. Supervised Fine-Tuning on 7.5K GSM8k training data.}
\label{app:self-sft-7.5k}
\begin{table}[h]
    \centering
    \small
    \caption{Comparison between SELF and Supervised Fine-Tuning on GSM8K. 
   A ``-'' symbol in the table indicates self-refinement was not conducted during inference because the model lacks the necessary self-refinement skill.}
     \vskip 0.15in
    \begin{center}
    \begin{small}
    \begin{adjustbox}{width=0.85\textwidth}
    \begin{tabular}{cccccc}
        \toprule
     \multirow{2}{*}{Direct Generation (\%)} & \multirow{2}{*}{Self-Refinement (\%)} &   \multirow{2}{*}{\(D_{\text{QA}}\) } &  \multirow{2}{*}{$D_{\text{meta}}$ } & \multicolumn{2}{c}{Self Evol.} \\
         \cmidrule(lr){5-6}
        &  & & & 1st  & 2nd  \\
        \midrule
        \midrule
        28.05 & - & $\checkmark$ & & & \\
        31.23 & 32.98 &  & $\checkmark$ & & \\
        35.43 & 36.22 &  & $\checkmark$ &  & \\
        \textbf{37.87} & \textbf{38.12} &  & $\checkmark$ & $\checkmark$ & $\checkmark$ \\
        \midrule
         35.70 & - & SFT  & (GSM8K training data) & & \\
        \bottomrule
    \end{tabular}
    \end{adjustbox}
    \end{small}
    \end{center}
    \label{tab:self-vs-sft}
\end{table}

When fine-tuned on the GSM8K 7.5k training set, the Vicuna model achieves an accuracy of 35.70\%, which is lower than the SELF method (37.87\%).

The experiments in~ \cref{tab:self-vs-sft} use 7.5k meta-skill data, ensuring a fair comparison with the supervised fine-tuned model. This approach differs from the one in~\cref{tab:main-table}, where only 3.5k meta-skill data are used.

\cref{tab:self-vs-sft} indicates that, with 7.5k unlabeled training prompts for the meta-skill learning corpus, Vicuna + $D_{\text{QA}}$ achieves 28.05\%. Post meta-skill learning, direct generation results improve to 31.23\%, further increasing to 32.98\% after self-refinement. Subsequent self-evolution rounds lead to performance gains, reaching 37.87\%(direct generation) and 38.12\%(self-refinement) in the second round, outperforming supervised fine-tuning (35.70\%).

\paragraph{Continuous Improvement of SELF vs. Supervised Fine-tuning:} SELF's primary advantage lies in its ability for continuous improvement and adaptation. In contrast to supervised fine-tuning, SELF doesn't rely on human or external LLM annotations (like GPT3.5/GPT4) for training data in self-evolution training.

\subsection{Scalability of SELF Framework}
\label{app:self-scalability}
To explore how SELF performs with different starting model qualities, we conduct experiments using the OpenLlama-3b model~\citep{openlm2023openllama}, a smaller LLM along with a stronger LLM, VicunaV1.5(finetuned from Llama2-7b)l~\citep{vicuna2023}, on the GSM8K dataset. 
This allows us to assess SELF’s adaptability to model quality. Experiments with SELF are based on the first round of self-evolution. 
The results are as follows:
\begin{table}[h]
    \centering
    \small
    \caption{Scalability of the SELF framework across different models.\label{tab:self-openllama-vicunav1.5}}
 \vskip 0.15in
    \begin{center}
    \begin{small}
    \begin{adjustbox}{width=0.75\textwidth}
    \begin{tabular}{l c c}
        \toprule
       {Model} & {Direct Generation (\%)} &{Self-Refinement (\%)} \\
        \midrule
        OpenLlama-3b & 2.04 & 1.01 \\
        OpenLlama-3b + $D_{\text{QA}}$ & 12.13 & 10.97 \\
        OpenLlama-3b + $D_{\text{QA}}$ + SELF & 15.32 & 15.78 \\
        \midrule
        Vicuna (Llama-7b) & 16.43 & 15.63 \\
        Vicuna + $D_{\text{QA}}$ & 24.49 & 24.44 \\
        Vicuna + $D_{\text{QA}}$ + SELF & 27.67 & 29.34 \\
        \midrule
        VicunaV1.5 (Llama2-7b) & 18.5 & 17.43 \\
        VicunaV1.5 + $D_{\text{QA}}$ & 26.04 & 25.48 \\
        VicunaV1.5 + $D_{\text{QA}}$ + SELF & \textbf{30.22} & \textbf{32.43} \\
        \bottomrule
    \end{tabular}
    \end{adjustbox}
    \end{small}
    \end{center}
\end{table}

 \paragraph{Applicability and Robustness of SELF Framework:} The average improvement of 17.32\% via direct generation and 16.87\% after self-refinement underscores the framework's scalability and efficacy. 
 It reveals a consistent positive impact of the SELF Framework across diverse models. 

 \paragraph{SELF Framework exhibits enhanced performance on more powerful models:}
 As shown in~\cref{tab:self-openllama-vicunav1.5}, applying SELF to VicunaV1.5 results in the most significant gains - 30.22\% in direct generation and 32.43\% after self-refinement, surpassing the performance on Vicuna and OpenLlama-3b. This indicates that the effectiveness of the SELF framework improves with the underlying model's capabilities.

 \subsection{Impact of Meta-Skill Corpus Quality}
 \label{app:meta-skill-quality}
We examine the influence of meta-skill learning quality on the self-evolution process with the following results:

\begin{table}[h]
    \centering
    \small
    \caption{Effect of meta-skill corpus quality on model performance using GPT-3.5-turbo and GPT4.}
\vskip 0.15in
    \begin{center}
    \begin{small}
    \begin{adjustbox}{width=0.8\textwidth}
    \begin{tabular}{l c c}
        \toprule
       \multirow{2}{*}{{Training Stage}} & {Direct Generation (\%)} & {Self-Refinement (\%)} \\
        & {(GPT-3.5-turbo/GPT4)} & {(GPT-3.5-turbo/GPT4)}\\
        \midrule
        Vicuna + \(D_{\text{meta}}\) & 24.84/25.39 (0.55$\uparrow$) & 25.22/28.28 (3.06$\uparrow$) \\
        Vicuna + \(D_{\text{meta}}\) + SELF Evol. & 25.11/27.67 (2.56$\uparrow$) & 25.47/29.34 (3.87$\uparrow$) \\
        \bottomrule
    \end{tabular}
    \end{adjustbox}
    \end{small}
    \end{center}
    \label{tab:self-meta-data-quality}
\end{table}

The presented \cref{tab:self-meta-data-quality} demonstrates the remarkable performance improvements achieved by using GPT-4 for generating the meta-skill corpus in our SELF framework, compared to using GPT-3.5-turbo. The table shows significant enhancements in both direct generation and self-refinement across training stages when GPT-4 is utilized. 
For instance, in the ``Vicuna + \(D_{\text{meta}}\)'' stage, direct generation performance increases from 24.84\% with GPT-3.5-turbo to 25.39\% with GPT-4, marking a gain of 0.55\%. Similarly, in the ``Vicuna + \(D_{\text{meta}}\) + SELF Evolution'' stage, the self-refinement result improves from 25.47\% with GPT-3.5-turbo to 29.34\% with GPT-4, showing an enhancement of 3.87\%.

This analysis highlights the significant impact of utilizing high-quality meta-skill training data on the performance of the Vicuna model within the SELF framework. The shift from GPT-3.5-turbo to GPT-4 for the generation of the meta-skill corpus leads to consistent improvements in both Direct Generation and Self-Refinement metrics.

\subsection{Single-Round vs. Iterative Self-Evolution Training}
\label{app:single-multiple}
Given an equal number of unlabeled prompts, we evaluate the effectiveness of training within a single-round versus iterative training. 
The former method uses a single model to self-curate training data from all available unlabeled prompts at once.
In contrast, the latter method involves dividing the unlabeled prompts into multiple parts.  
For the iterative approach, the model is initially trained on a portion of the unlabeled prompts and self-curated labels. Following this, the trained model is employed to create new training data based on previously unused prompts. As described in our main text, we divide the unlabeled prompts into three parts, enabling the model to undergo three iterative rounds of self-evolution.

\begin{table}[h]
    \centering
    \small
    \caption{Comparison of single-round training and iterative training.}
\vskip 0.15in
    \begin{center}
    \begin{small}
    \begin{adjustbox}{width=0.7\textwidth}
    \begin{tabular}{l c c}
        \toprule
       {Training Method} & {Direct Generation (\%)} & {Self-Refinement (\%)} \\
        \midrule
        SELF (Single-Round) & 28.40 & 30.55 \\
        SELF (Iterative) & 29.64 & 31.31 \\
        \bottomrule
    \end{tabular}
    \end{adjustbox}
    \end{small}
    \end{center}
    \label{tab:single-multiple-self-evolution}
\end{table}

\cref{tab:single-multiple-self-evolution}  shows that in the ``Single-Round'' training, the performance is 28.40\% for direct generation and 30.55\% for self-refinement. In contrast, the iterative approach yields higher scores of 29.64\% for direct generation and 31.31\% for self-refinement.
\paragraph{Advantages of Iterative Training:} Iterative training benefits from the enhanced capabilities of LLMs in subsequent rounds, which generate higher-quality training data and lead to improved test performance.

\end{document}